\documentclass[authoryear,preprint,review,10pt]{elsarticle}
\usepackage{etoolbox}
\usepackage{graphicx}
\usepackage{amssymb}
\usepackage{amsmath}
\usepackage{amsthm}
\usepackage{booktabs}
\usepackage{siunitx}
\usepackage{multirow}
\usepackage{threeparttable}
\usepackage{makecell}
\usepackage{tabularx}
\usepackage{longtable}
\usepackage{lscape}
\usepackage{rotating}
\usepackage{float}
\usepackage{url}

\usepackage{algorithm}
\usepackage{algpseudocode}

\algrenewcommand\algorithmicrequire{\textbf{Input:}}
\algrenewcommand\algorithmicensure{\textbf{Output:}}

\usepackage{array}
\usepackage{tabularx}
\newcolumntype{Y}{>{\raggedright\arraybackslash}X}

\begin{document}
\begin{frontmatter}

\title{Implicit neural representations as a coordinate-based framework for continuous environmental field reconstruction from sparse ecological observations}

\author[1]{Agnieszka Pregowska}\corref{cor1}
\ead{aprego@ippt.pan.pl}

\author[2,3]{Hazem M. Kalaji}

\cortext[cor1]{Corresponding author}

\affiliation[1]{
  organization={Institute of Fundamental Technological Research, Polish Academy of Sciences},
  addressline={Pawinskiego 5B},
  city={Warsaw},
  postcode={02-106},
  country={Poland}
}

\affiliation[2]{
  organization={Department of Plant Physiology, Institute of Biology, Warsaw University of Life Science},
  addressline={159 Nowoursynowska St},
  city={Warsaw},
  postcode={02-776},
  country={Poland}
}

\affiliation[3]{
  organization={Institute of Technology and Life Sciences Falenty},
  addressline={Al. Hrabska 3},
  city={Raszyn},
  postcode={05-090},
  country={Poland}
}

\begin{abstract}
Reconstructing continuous environmental fields from sparse and irregular observations remains a central challenge in environmental modelling and biodiversity informatics. Many ecological datasets are heterogeneous in space and time, making grid-based approaches difficult to scale or
generalise across domains.
Here, we evaluate implicit neural representations (INRs) as a
coordinate-based modelling framework for learning continuous spatial and spatio-temporal fields directly from coordinate inputs. We analyse their behaviour across three representative modelling scenarios: species distribution reconstruction, phenological dynamics, and morphological segmentation derived from open biodiversity data. Beyond predictive performance, we examine interpolation behaviour,
spatial coherence, and computational characteristics relevant for environmental modelling workflows, including scalability, resolution-independent querying, and architectural inductive bias. Results show that neural fields provide stable continuous representations with predictable computational cost, complementing classical smoothers and tree-based approaches.
These findings position coordinate-based neural fields as a flexible representation layer that can be integrated into environmental modelling pipelines and exploratory analysis frameworks for large, irregularly sampled datasets.
\end{abstract}


\end{frontmatter}

\section{Introduction}

Understanding how plant species are distributed in space, how their seasonal rhythms unfold in time, and how their morphology varies across environments lies at the core of plant ecology and biodiversity science. Species distributions emerge from interactions between physiological tolerances, dispersal processes, and environmental gradients, and shifts in these distributions are among the most widely documented biological responses to climate change \citep{ParmesanYohe2003,Thuiller2005,Sillero2021}. Similarly, plant phenology, including flowering and leaf-out timing, responds sensitively to temperature and photoperiod, making phenological observations key indicators of environmental change \citep{MillerRushingPrimack2008,Piao2019}. Morphological traits, such as leaf shape and size, further reflect both adaptive and plastic responses to environmental constraints and can vary continuously across spatial gradients \citep{Violle2007,PerezHarguindeguy2013}.

Physiological responses of plants to environmental gradients manifest as continuous changes in growth, development, phenology, and morphology. Temperature, moisture availability, photoperiod, soil properties, and biotic interactions shape species’ ranges and seasonal dynamics through gradual transitions rather than sharp boundaries. However, most contemporary biodiversity datasets, particularly those derived from global citizen-science platforms, sample these continuous processes irregularly and unevenly in space and time. Observations are often sparse, clustered, and biased by uneven sampling effort, creating a fundamental mismatch between the continuity of ecological processes and the discreteness of available data \citep{isaac2014,ElithLaethwick2009}.

In environmental modelling, space and time often function as primary environmental coordinates, even in the absence of explicit covariates such as climate or land-use variables. Reconstructing continuous environmental fields from irregularly sampled observations is therefore a fundamental modelling task, underpinning applications ranging from spatial interpolation and gap-filling to exploratory pattern analysis. Methods that decouple field representation from fixed grids and allow resolution-independent querying are particularly valuable in this context, as they support scalable inference across heterogeneous spatial and temporal domains.

A wide range of statistical and machine-learning approaches has been developed to address these challenges. In species distribution modelling (SDM), commonly used baselines include regularized presence-only models such as MaxEnt \citep{Phillips2006}, and tree-based ensembles including boosted regression trees and random forests \citep{ElithLaethwick2009,Breiman2001}. Bayesian and spatially explicit approaches, such as Gaussian process models and related geostatistical formulations, provide continuous spatial predictions together with uncertainty estimates \citep{RasmussenWilliams2006}, while scalable Bayesian spatial models based on stochastic partial differential equations (SPDE) and integrated nested Laplace approximation (INLA) have been increasingly adopted in ecological applications \citep{Rue2009,Lindgren2011}. More recently, deep learning methods have been explored for ecological prediction tasks, particularly where heterogeneous covariates or large volumes of data are available \citep{Christin2019}. Each of these approaches involves trade-offs between interpretability, scalability, and computational cost, especially when dense spatial or spatiotemporal predictions are required.

Implicit Neural Representations (INRs), also referred to as neural fields or coordinate-based neural networks, have recently emerged as a flexible framework for learning continuous functions directly from coordinate inputs \citep{park2019,mescheder2019}. In this paradigm, a neural network represents a signal as a continuous function of its spatial or spatiotemporal coordinates, rather than as values on a fixed grid. Extensions such as sinusoidal activation functions (SIREN) and Fourier-feature encodings substantially enhance the ability of these models to represent fine-scale structure and high-frequency variation \citep{sitzmann2020,tancik2020}. INRs have achieved notable success in computer vision and graphics, where they enable high-resolution reconstruction of geometry, occupancy, and radiance fields from sparse observations \citep{mildenhall2022}. More broadly, INRs belong to a wider family of continuous machine-learning models for scientific data, alongside Gaussian processes and geostatistical models that offer uncertainty quantification \citep{RasmussenWilliams2006}, spline-based smoothers such as GAMs \citep{Wood2017}, and scalable Bayesian spatial approaches (e.g., SPDE/INLA) \citep{Rue2009,Lindgren2011}. Graph-based and structured probabilistic models can further encode spatial dependencies and interactions, while ensemble and composite methods often improve robustness under dataset shift and heterogeneous noise. In this context, INRs can be viewed as a complementary function class that emphasizes expressive coordinate-to-signal mappings and fast dense querying after training, but typically requires additional machinery to obtain calibrated uncertainty.

From an ecological modelling perspective, INRs occupy a position that is complementary to both tree-based machine-learning models and classical spatial smoothers. Compared to tree ensembles trained on raw coordinates, INRs impose a smooth functional inductive bias that supports interpolation between observations rather than memorization of sampled locations. Compared to Gaussian processes and spline-based approaches, INRs replace explicit kernel or basis-function design with a learnable parametric representation that can be efficiently evaluated on dense query grids once trained, although standard INRs do not provide calibrated uncertainty estimates by default. This makes INRs particularly attractive for tasks focused on reconstructing continuous spatial or spatiotemporal fields from irregularly sampled data, while highlighting the need for careful evaluation and interpretation.

Despite these properties, the potential of INRs for ecological applications has so far received limited systematic attention. In particular, their behaviour across different ecological data modalities, such as species occurrences, phenological observations, and morphological traits, and their relationship to commonly used ecological baselines remain underexplored. Moreover, coordinate-only models are susceptible to sampling bias and spatial clustering, making transparent evaluation and cautious interpretation essential when assessing apparent predictive performance.

Environmental modelling increasingly relies on reusable computational representations that can integrate heterogeneous data sources within reproducible software workflows. In this context, we introduce implicit
neural representations as a coordinate-based modelling abstraction that can function as a modular component within environmental modelling software pipelines.

In this study, we investigate implicit neural representations as a general-purpose modelling abstraction for continuous environmental fields learned from irregular coordinate samples. Rather than focusing on a single ecological application, we analyse how coordinate-based
neural fields behave across multiple environmental data modalities and evaluate their computational and representational properties. We compare several INR architectures against classical coordinate baselines and assess predictive performance, spatial structure of
reconstructed fields, and computational scalability. Our goal is to situate neural fields within the broader landscape of environmental modelling tools, highlighting their potential role as reusable representation components within modelling workflows and software pipelines. Importantly, our goal is not to propose a new ecological prediction model but to evaluate coordinate-based neural fields as a reusable computational abstraction that can be integrated into environmental modelling workflows and software systems.

\section{Datasets}

We used four datasets spanning three plant-ecology tasks: species distribution modelling (GBIF), phenology reconstruction (iNaturalist in two coordinate formulations: iNat-2D and iNat-3D), and leaf-morphology segmentation (Leafsnap) (Table~\ref{tab:datasets}).
These datasets were selected to represent common data modalities in contemporary biodiversity informatics: point occurrences with presence-background sampling, opportunistic phenological event records with strong spatiotemporal bias, and high-resolution image-derived masks enabling dense spatial sampling. Such diversity allows us to evaluate model behaviour under heterogeneous sampling regimes and input domains that are typical for open biodiversity resources \citep{ElithLaethwick2009,isaac2014}.

\subsection{GBIF (species distribution; presence-background)}
The GBIF dataset consists of georeferenced occurrence records for Quercus robur and a set of background samples represented by longitude and latitude coordinates. GBIF provides globally aggregated biodiversity observations that are widely used in correlative SDM workflows \citep{gbif2023,Sillero2021}. Because presence-only occurrence data require a contrast class for supervised learning, we follow the standard presence-background setup commonly used in SDM, where background points approximate the available environmental/geographic space \citep{ElithLaethwick2009,Phillips2006}. This dataset therefore serves as a canonical benchmark for coordinate-only niche modelling under sparse predictors and heterogeneous observation density.

\subsection{iNaturalist iNat-2D and iNat-3D (phenology; opportunistic flowering events)}
The iNaturalist-based phenology datasets contain georeferenced observations annotated as flowering events. We consider two variants: iNat-2D, where models learn spatial flowering patterns from (lon, lat) coordinates alone, and iNat-3D, where the day of year (DOY) is included to learn a spatiotemporal flowering field in (lon, lat, DOY). iNaturalist is a major citizen-science platform that provides large volumes of opportunistic observations; however, such data are affected by strong spatial clustering, temporal sampling variation, and observer-effort bias \citep{inaturalist2025,isaac2014,cecco2021}. These characteristics make iNat datasets a challenging testbed for environmental field reconstruction, particularly under coordinate-only inputs where sampling bias can be confounded with ecological signal.

\subsection{Leafsnap (leaf morphology; segmentation masks)}
The Leafsnap dataset provides leaf images together with segmentation masks, enabling pixel-level learning of leaf/background structure \citep{kumar2012}. In our setup, we treat the mask as a dense set of labelled spatial samples in image coordinates (x, y), which allows us to evaluate how different INR bases represent sharp boundaries and fine-scale morphological detail. Because the effective number of training samples is determined by the number of sampled pixels rather than the number of images, this dataset represents a complementary regime to point-occurrence data: it is locally dense, high-resolution, and dominated by boundary geometry rather than broad-scale spatial gradients \citep{kumar2012}.

\subsection{Quantitative dataset properties and sampling structure}

Table~\ref{tab:datasets} summarizes the main characteristics of the datasets used in this study.
For the GBIF species distribution task, we used approximately 3{,}000 georeferenced presence
records for \textit{Quercus robur} together with a matched set of background points after
filtering and deduplication. Presence records are strongly clustered in Europe, reflecting
known sampling biases related to accessibility and observer effort.

For the iNaturalist phenology task, we considered two dataset variants derived from the same source domain: iNat-2D and iNat-3D. Both variants contain approximately 50{,}000 records in a presence-background design, with iNat-3D additionally including day-of-year (DOY) as a third coordinate. The ratio of flowering observations to background points was approximately 1:1.
Observations are highly uneven in both space and time, with strong concentration in temperate regions and during peak flowering periods, consistent with well-documented biases in citizen-science phenological data.

The Leafsnap dataset represents a fundamentally different sampling regime, consisting of densely sampled image-plane coordinates rather than sparse geographic points. After subsampling segmentation masks, approximately $2\times10^5$ pixel-level observations were used, with substantial class imbalance between leaf and background pixels. This dataset emphasizes
local geometric structure and sharp boundaries rather than large-scale spatial gradients.

Together, these quantitative properties demonstrate that the datasets span complementary regimes of sparsity, bias, dimensionality, and spatial structure, providing a stringent testbed for coordinate-based models under conditions typical of open ecological data.

\subsection{Rationale and implications}
Together, these datasets span a wide spectrum of test conditions: from sparse point occurrences (GBIF) to large-scale, biased event records (iNaturalist) and dense image-plane sampling (Leafsnap). This diversity enables a more comprehensive evaluation of model generality, stability, and reconstruction behaviour across ecological domains. Importantly, we emphasize that coordinate-only learning can be particularly sensitive to sampling bias and spatial leakage, especially in citizen-science datasets \citep{isaac2014,cecco2021}. We therefore complement standard predictive metrics with spatial summary measures characterizing the structure and coherence of reconstructed probability fields, and we discuss limitations and evaluation considerations for ecological interpretation \citep{ElithLaethwick2009,Sillero2021}.

\begin{table}[H]
\centering
\tiny
\caption{Summary of plant ecology datasets used in this study.}
\label{tab:datasets}
\begin{threeparttable}
\begin{tabular}{llllr}
\toprule
Dataset & Phenomenon & Target variable & Records & Input dims \\
\midrule
GBIF      & Species distribution        & presence / background         & 3\,000   & lon, lat (2D) \\
iNat-2D   & Phenology (spatial)         & flowering / background        & 50\,000  & lon, lat (2D) \\
iNat-3D   & Phenology (spatiotemporal)  & flowering / background        & 50\,000  & lon, lat, DOY (3D) \\
Leafsnap  & Leaf morphology             & pixel mask (leaf / background)& 200\,000\tnote{a} & x, y (2D) \\
\bottomrule
\end{tabular}
\begin{tablenotes}
\item[a] Number of sampled segmentation mask points, not number of images.
\item[] iNat-2D and iNat-3D are two coordinate formulations of the same phenology source domain.
\end{tablenotes}
\end{threeparttable}
\end{table}

\subsection{Integration across ecological data types}
In this study, integration refers to the use of a single coordinate-based modelling framework applied consistently across distinct ecological tasks, rather than to a joint mechanistic model linking species distributions, phenology, and morphology. By using the same functional representation, training protocol, and evaluation strategy across datasets, we demonstrate how INRs can serve as a unifying computational abstraction for heterogeneous ecological data modalities. Achieving true biological integration of these processes remains an important challenge for future work.

\subsection{Data sources, preprocessing, and sampling}
All datasets were obtained from public repositories and were processed using a unified workflow to enable coordinate-only learning. For each dataset, we removed records with missing coordinates, filtered obvious spatial outliers (invalid longitude/latitude ranges), removed duplicate coordinate pairs, and standardized coordinate units and ranges. For citizen-science datasets (GBIF and iNaturalist), we additionally acknowledge heterogeneous observation effort and reporting biases that can induce strong spatial clustering and temporal sampling artefacts \citep{isaac2014,cecco2021,ElithLaethwick2009}.

\subsection{Background / pseudo-absence construction}
For GBIF and iNaturalist tasks, we used a presence-background design, where background points represent the available geographic (or spatiotemporal) domain rather than confirmed absences \citep{Phillips2006,ElithLaethwick2009}. Background points were sampled uniformly within the spatial bounding box of the filtered presences (and uniformly over DOY for the 3D phenology case), unless stated otherwise. This design follows standard SDM practice while making explicit that predicted probabilities are conditional on the chosen background definition \citep{Sillero2021}.

\subsection{Coordinate normalization}
All coordinate inputs were linearly rescaled to $[-1,1]$ per dimension based on the training domain to stabilize optimization and to make architectures comparable across input spaces.

\subsection{Domain characteristics}
The four datasets represent complementary sampling regimes: sparse and heterogeneous point occurrences (GBIF), large-scale and biased opportunistic event observations with strong spatial and temporal clustering (iNat-2D and iNat-3D), and locally dense, high-resolution image-plane samples dominated by boundary geometry (Leafsnap). This diversity is essential for evaluating how different function classes behave under irregular sampling and coordinate-only inputs \citep{isaac2014,ElithLaethwick2009}.

\section{Modelling procedure}

\subsection{Problem formulation}
For GBIF and iNaturalist, we model a binary label $y \in \{0,1\}$ (presence/event vs. background) as a function of coordinates $\mathbf{x} \in \mathbb{R}^d$, with $d=2$ for spatial tasks and $d=3$ for spatiotemporal phenology. For Leafsnap, we model a binary pixel label (leaf vs. background) as a function of image-plane coordinates $\mathbf{x}=(x,y)$. Model outputs are logits transformed with a sigmoid to yield $p_\theta(\mathbf{x}) \in (0,1)$. Because all species distribution and phenology tasks are formulated in a presence-background setting, the predicted values should be interpreted as relative occurrence or event-intensity scores conditional on the chosen background sampling scheme, rather than as absolute probabilities of presence. As in standard correlative SDM practice, these scores depend on the spatial and temporal extent of the background and the prevalence of presences in the training data \citep{Phillips2006,ElithLaethwick2009,Sillero2021}.

In this formulation, spatial and temporal coordinates are treated as environmental dimensions that define the domain of the reconstructed field. Model outputs therefore represent continuous intensity or probability surfaces over the environmental coordinate space, rather than mechanistic ecological processes.

\subsection{Train/validation/test splits and spatial robustness}
\label{sec:splits}

Because biodiversity observations are spatially clustered and autocorrelated, random train-test splits can lead to spatial leakage and inflated predictive scores. To quantify this effect, we report results under two protocols. We randomly split records into train and test sets (80/20). This protocol reflects conventional machine-learning evaluation but does not control for spatial dependence. For the 2D datasets (GBIF and iNat-2D), we partitioned the longitude-latitude domain into non-overlapping spatial blocks of size $5^\circ \times 5^\circ$ and assigned entire blocks to either train or test (approximately 80/20 of blocks).
For the 3D phenology dataset (iNat-3D), we additionally blocked along day-of-year using 30-day bins, yielding spatio-temporal blocks in $(\mathrm{lon}, \mathrm{lat}, \mathrm{DOY})$.

These blocked protocols reduce spatial leakage by ensuring that test points are geographically (and temporally) separated from training data. We interpret blocked-split
performance as a more conservative proxy for spatial generalization, while random-split performance primarily reflects interpolation within clustered sampling footprints.

\subsection{Models}
\paragraph{INR architectures}
We evaluated four INR variants: SIREN (sine activations) \citep{sitzmann2020}, Fourier-feature MLPs \citep{tancik2020}, ReLU MLPs \citep{rahaman2018}, and an RBF-based INR for Leafsnap to better represent sharp boundaries. All INR models share the same depth and width to isolate the impact of encoding/basis functions.

\subsection{Baselines}
Random Forest classifiers were trained on the same coordinate-only inputs as the INR
models (500 trees, maximum depth 20, minimum 5 samples per leaf, bootstrap enabled).
As a non-smooth, high-capacity coordinate baseline, Random Forests provide a useful contrast
to INRs by highlighting localization and spatial fragmentation effects under random splits.

To benchmark INRs against a classical continuous smoother, we implemented a GAM-like baseline using tensor-product B-spline bases over the coordinate inputs
and a logistic regression model with $\ell_2$ regularization.
Specifically, for $d$ input dimensions we construct a tensor-product spline design
matrix and fit a penalized logistic model, where the regularization strength acts
as a smoothing proxy. Unlike tree ensembles trained on raw coordinates, this
baseline yields smooth probability fields and provides a principled non-neural
reference for coordinate-only reconstruction.
All spline bases were fitted on the training domain and re-used for test data,
with fixed lower/upper bounds to prevent out-of-range knot issues under blocked splits. Bounds were fixed using the overall coordinate range of the dataset (prior to splitting) to avoid extrapolation failures; this affects only basis evaluation and does not use test-set labels.

\subsection{Optimization and hyperparameters}

All INR models were trained using the Adam optimizer with a learning rate of $10^{-3}$ and a 
batch size of 4096. Models were trained for a maximum of 10 epochs for the GBIF and iNaturalist
tasks and 8 epochs for the Leafsnap morphology task, with early stopping based on validation loss (patience = 3 epochs). Each experiment was trained with a fixed random initialization. Random Forest baselines were trained with 500 trees, a maximum depth of 20, and a minimum of 5 samples per leaf. Bootstrap sampling was enabled. Hyperparameters were fixed across datasets to ensure comparability and to avoid dataset-specific tuning advantages. All coordinate inputs were normalized to the range $[-1,1]$ per dimension based on the training  domain. Full architectural details are reported in Supplementary Table \ref{tab:s1}.

\subsection{Metrics}
We report standard classification metrics (ROC AUC, PR AUC, log loss, Brier score), which quantify the ability of models to discriminate presences from background points under the given sampling design. These metrics evaluate relative ranking performance rather than calibrated probabilities of occurrence. Because these are not mechanistic ecological indicators, we explicitly treat them as descriptive summaries of reconstructed fields rather than ecological process metrics \citep{ElithLaethwick2009,Sillero2021}. For morphology, we report pixel-level metrics (F1, IoU, Dice) and discuss boundary behaviour qualitatively.

\section{Implicit Neural Representations}
\subsection{General formulation}
We model all ecological signals in this study with INRs, which are continuous coordinate-based neural functions \citep{park2019,mescheder2019,sitzmann2020,tancik2020}. Formally, an INR is a parametric function
\[
f_\theta : \mathbb{R}^d \to \mathbb{R}^c,
\]
implemented by a neural network with parameters~$\theta$, which maps a coordinate vector $\mathbf{x} \in \mathbb{R}^d$ to one or more ecological quantities of interest $\mathbf{y} \in \mathbb{R}^c$. INRs provide a flexible and differentiable way to model spatial and temporal fields. They can capture both smooth gradients and high-frequency ecological structure, without the need for a predefined grid.
This continuous formulation has recently gained popularity in tasks that require high-resolution reconstructions from sparse data \cite{mildenhall2022,verbin2025}. It is therefore well suited to ecological modelling, where irregular sampling is common \cite{isaac2014}.

In addition, input coordinates encode either geographical location, using longitude and latitude, for species distribution data. They can also encode spatiotemporal position, using longitude, latitude and day of year, for phenological observations. Moreover, they can encode the location of the image-plane, using pixel coordinates, for morphological segmentation tasks. These three coordinate systems correspond, respectively, to the GBIF dataset \cite{gbif2023}, the iNaturalist phenology observations \cite{bonne2019,inaturalist2025}, and the Leafsnap leaf-morphology dataset \cite{kumar2012}, as summarized in Table~\ref{tab:datasets}. Once trained, $f_\theta$ can be queried anywhere within the spatial or temporal domain, providing resolution-independent access to the underlying ecological surface. This allows us to compute descriptive summaries of reconstructed probability fields, such as continuous occurrence or event-probability surfaces, geometric range descriptors (e.g., extent of occurrence), spatially aggregated phenological patterns, or leaf-shape reconstructions. These summaries describe properties of the learned fields rather than mechanistic ecological processes.
The differentiability of $f_\theta$ with respect to inputs enables gradient-based sensitivity analysis in principle, although such analyses are not explored in this study  \cite{briscoe2023,brun2022}. 

\subsection{Architectures evaluated}
\begin{figure}[t]
\centering
\includegraphics[width=\textwidth]{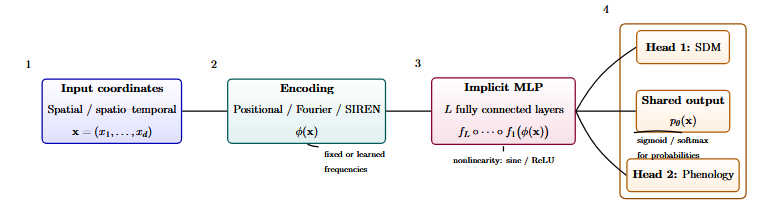}
\label{fig:inr_architecture}
\end{figure}

We compare several INR architectures that differ in their ability to encode high-frequency variation and represent complex ecological patterns, as we show in Figure
 \ref{fig:inr_architecture}. First, we evaluate sine-activated SIREN networks \cite{sitzmann2020}, which have demonstrated strong performance in reconstructing fine-scale spatial structure due to their inherent ability to represent high-frequency components. Second, we investigate MLPs augmented with Fourier feature encodings \cite{tancik2020}, which randomize sinusoidal projections of the input coordinates to increase spectral expressiveness. Third, we include standard ReLU-based MLPs as a baseline architecture commonly used in coordinate-based modelling but known to struggle with high-frequency detail \cite{rahaman2018}. All models share the same supervised learning formulation. Given coordinate-label pairs $\{(\mathbf{x}_i, \mathbf{y}_i)\}_{i=1}^N$ sampled from the GBIF, iNaturalist, and Leafsnap datasets, we optimize parameters $\theta$ to minimize a task-specific loss that reflects the ecological target. We use binary cross-entropy for distribution and phenology tasks, and a BCE-type loss for leaf segmentation. This unified function-learning framework enables a consistent treatment of species distribution modelling \citep{ElithLaethwick2009}, large-scale phenological analyses based on opportunistic observations \citep{millerrushing2008}, and
image-based species identification and leaf trait extraction \citep{kumar2012}. It also facilitates direct comparison of different coordinate encodings and architectural choices.

Predictive performance was quantified using ROC AUC, PR AUC, Brier score, and log loss. In addition, we computed spatial summary metrics describing the structure of predicted probability fields, including predicted area above a fixed threshold, extent of occurrence (EOO), and hit rates within the highest-probability regions. These metrics characterize geometric and spatial properties of the reconstructed fields rather than ecological mechanisms per se.

\subsection{Ecological context and advantages}
Ecological data are typically sparse, noisy, and unevenly distributed in space and time. These issues are especially pronounced when data come from opportunistic or citizen-science sources. Species distribution models (SDMs) must reconcile irregular occurrence records with environmental predictors that are defined on heterogeneous spatial and temporal grids \cite{ElithLaethwick2009}. Similarly, phenological datasets often combine long-term, site-based observations with opportunistic records and remote-sensing products. This combination leads to strong scale mismatches between ecological processes and available data \cite{millerrushing2008}. Citizen-science platforms such as iNaturalist further amplify these challenges by introducing spatial and temporal biases in observation effort \cite{isaac2014}. In this context, methods that can directly learn continuous ecological surfaces from irregular samples, without committing to a fixed discretisation, are particularly attractive.

Implicit Neural Representations address this challenge by representing ecological signals as continuous functions over coordinate space, rather than as discrete values on a predefined raster. INRs have been shown to accurately reconstruct complex high-frequency signals such as geometry, occupancy fields, or radiance, even from sparse observations \citep{park2019,mescheder2019,sitzmann2020,tancik2020,mildenhall2022}. Their ability to represent detailed structure while remaining parameter-efficient is especially important for ecological applications. In these settings, dense gridded observations are rarely available, yet high-resolution predictions are often required for management and conservation.

From an ecological perspective, the main advantage of INRs is that they decouple the representation of ecological processes from any specific grid resolution or projection. Once trained, an INR can be queried at arbitrary spatial or temporal resolution, yielding continuous suitability surfaces, range boundaries, or phenological profiles from the same underlying function. This is analogous to geostatistical or Gaussian-process approaches, which also model environmental fields from irregular data, but INRs scale more readily to large datasets and high-dimensional inputs. Because INRs are fully differentiable with respect to both inputs and parameters, they also support gradient-based sensitivity analyses and integration into larger end-to-end learning pipelines, including hybrid mechanistic–statistical models.

INRs provide a unifying framework across ecological data types.
The same coordinate-based formulation can be applied to point-level species distributions, spatiotemporal phenology, and high-resolution image-based morphology, simply by changing the input coordinates and output variable. This helps to bridge traditionally separate modelling traditions-raster-based SDMs, time-series phenology models, and image-based trait extraction, within a single framework, and enables direct comparison of how different architectures and coordinate encodings capture ecological structure across scales.

The computational workflow used in this study is summarized in the Supplementary Material (Algorithms S1-S5). These algorithms describe
dataset construction, coordinate normalization, training of implicit neural representations, continuous field reconstruction, and evaluation
under random and blocked protocols.

\section{Results}
We first evaluate the performance of implicit neural representations in the three ecological tasks of species distribution, phenology, and morphology. Across all datasets, INR models achieve competitive predictive accuracy compared to standard baselines, while exhibiting markedly different interpolation behaviour and spatial field structure.

Throughout the Results, we focus on interpolation behaviour, spatial coherence, and computational characteristics of the reconstructed fields, rather than on mechanistic ecological interpretation of predicted values.

We then analyze how different coordinate encodings and architectural choices (SIREN, Fourier features, ReLU-MLPs) affect model performance and the ability to capture fine-scale ecological structure. Unless stated otherwise, Table~\ref{tab:predictive} reports results under a conventional random 80/20 train-test split. We then quantify spatial (and spatio-temporal) leakage using blocked evaluation protocols (Section~\ref{sec:splits}).

INR models achieved high discriminative performance across datasets, with SIREN and Fourier
variants generally outperforming the ReLU MLP, whereas Random Forests reached near-perfect
scores under random splits consistent with spatial leakage in clustered observations
(Table~\ref{tab:predictive}). All classification metrics (ROC AUC, PR AUC, LogLoss, Brier score, accuracy and F1 at a threshold of 0.5, and ECE) were calculated on presence-background training sets, and the reported confidence intervals are bootstrap-estimated. The perfect ROC AUC and PR AUC values obtained by the Random Forest under random train-test splits reflect strong spatial localization around sampled coordinates rather than genuine spatial generalization. When presences and background points are geographically clustered, random splitting induces substantial spatial leakage, allowing tree-based models to effectively memorize coordinate partitions. This behaviour is well documented in SDM studies using geographically biased data and highlights the limitations of random splits for evaluating coordinate-based models \citep{ElithLaethwick2009,isaac2014}.

\begin{table}[t]
\centering
\scriptsize
\caption{Predictive performance across datasets under a random 80/20 split. Metrics include 95\% bootstrap confidence intervals where applicable. Note that thresholded metrics (Acc@0.5, F1@0.5) can be misleading in presence--background settings; models may assign probabilities mostly below 0.5, yielding high accuracy but F1$\approx 0$ despite strong ranking performance (ROC/PR AUC).}

\label{tab:predictive}
\begin{threeparttable}
\begin{tabular}{llrrrrrrr}
\toprule
Dataset & Model & ROC AUC (95\% CI) & PR AUC (95\% CI) & LogLoss & Brier & Acc@0.5 & F1@0.5 & ECE \\
\midrule
GBIF   & SIREN   & 0.995 [0.995, 0.996] & 0.989 [0.988, 0.990] & 0.080 & 0.021 & 0.973 & 0.964 & 0.010 \\
GBIF   & Fourier & 0.995 [0.994, 0.995] & 0.989 [0.988, 0.990] & 0.072 & 0.018 & 0.977 & 0.970 & 0.005 \\
GBIF   & ReLU    & 0.986 [0.986, 0.987] & 0.969 [0.966, 0.971] & 0.128 & 0.034 & 0.958 & 0.945 & 0.016 \\
GBIF   & RF      & 1.000 [1.000, 1.000] & 1.000 [1.000, 1.000] & 0.010 & 0.002 & 1.000 & 1.000 & 0.009 \\
\midrule
iNat2D & SIREN   & 0.948 [0.943, 0.954] & 0.560 [0.536, 0.584] & 0.083 & 0.023 & 0.970 & 0.471 & 0.007 \\
iNat2D & Fourier & 0.938 [0.932, 0.944] & 0.547 [0.521, 0.572] & 0.094 & 0.026 & 0.962 & 0.000 & 0.014 \\
iNat2D & ReLU    & 0.768 [0.759, 0.776] & 0.205 [0.186, 0.223] & 0.146 & 0.036 & 0.962 & 0.000 & 0.009 \\
iNat2D & RF      & 1.000 [1.000, 1.000] & 1.000 [1.000, 1.000] & 0.008 & 0.001 & 1.000 & 1.000 & 0.007 \\
\midrule
iNat3D & SIREN   & 0.978 [0.973, 0.982] & 0.812 [0.795, 0.828] & 0.056 & 0.014 & 0.982 & 0.750 & 0.012 \\
iNat3D & Fourier & 0.912 [0.905, 0.920] & 0.652 [0.633, 0.671] & 0.105 & 0.028 & 0.962 & 0.000 & 0.020 \\
iNat3D & ReLU    & 0.862 [0.855, 0.868] & 0.324 [0.304, 0.345] & 0.129 & 0.034 & 0.962 & 0.000 & 0.011 \\
iNat3D & RF      & 1.000 [1.000, 1.000] & 1.000 [1.000, 1.000] & 0.005 & 0.001 & 1.000 & 1.000 & 0.005 \\
\bottomrule
\end{tabular}
\end{threeparttable}
\end{table}

For the phenological tasks (iNat2D and iNat3D), performance gaps between INR variants are more pronounced: SIREN consistently delivers the highest predictive accuracy and stability, Fourier-based models perform at an intermediate level, and ReLU networks struggle with the heterogeneous and irregular data structure. Random forests again achieve near-perfect metrics under random splits, consistent with strong localization around sampled coordinates and potential spatial leakage when training and test points are geographically close. Taken together, these results show that INRs can achieve high predictive quality while avoiding artefacts that are characteristic of tree-based models.

\begin{table}[t]
\centering
\scriptsize
\caption{GAM-like tensor-spline baseline under random vs blocked evaluation protocols. Blocked splits enforce spatial (and spatio-temporal) separation to reduce leakage. Metrics are reported on the test set.}
\label{tab:gam_blocked}
\begin{tabular}{llrrrrr}
\toprule
Dataset & Split & ROC AUC & PR AUC & LogLoss & Brier & ECE \\
\midrule
GBIF   & blocked & 0.975 & 0.952 & 0.223 & 0.067 & 0.074 \\
GBIF   & random  & 0.991 & 0.990 & 0.119 & 0.031 & 0.034 \\
\midrule
iNat-2D & blocked & 0.884 & 0.843 & 0.394 & 0.120 & 0.072 \\
iNat-2D & random  & 0.944 & 0.950 & 0.317 & 0.096 & 0.043 \\
\midrule
iNat-3D & blocked & 0.940 & 0.891 & 0.289 & 0.087 & 0.060 \\
iNat-3D & random  & 0.964 & 0.967 & 0.249 & 0.074 & 0.063 \\
\bottomrule
\end{tabular}
\end{table}
For a smooth non-neural reference, we report the GAM-like tensor-spline baseline under both random and blocked protocols in Table~\ref{tab:gam_blocked}.
The effect is strongest for iNaturalist phenology, where opportunistic observations exhibit pronounced spatial aggregation.
These results support interpreting random-split scores primarily as within-footprint interpolation performance,
and using blocked splits as a more conservative estimate of spatial (and spatio-temporal) generalization.

Random splits can substantially overestimate performance in spatially clustered occurrence data due to spatial leakage. To obtain a more conservative estimate of spatial generalization, we therefore evaluated coordinate-only models under blocked cross-validation, where test samples are drawn from geographic blocks disjoint from the training blocks (and, for the 3D phenology case,
from disjoint spatiotemporal blocks).

Under blocked evaluation, all models exhibit lower discriminative performance compared to random splits, consistent with reduced spatial autocorrelation between train and test samples. Nevertheless, implicit neural representations retain coherent spatial reconstructions and maintain competitive ranking ability, whereas highly localized coordinate baselines degrade
more strongly. Detailed blocked-split metrics for GBIF and iNaturalist are reported in Table~\ref{tab:blocked_metrics}.

Spatial field summary metrics indicate that INRs produce broader and more spatially coherent high-probability regions than RF, whose predictions remain tightly concentrated around observed locations (Table~\ref{tab:eco}). Among the analyzed values are: area predicted as high probability, namely, area above 0.5, Extent of Occurrence (EOO), and two spatial accuracy indices Hit@1\% and Hit@5\%. For the GBIF data, all INR models produce almost identical large-scale spatial patterns, namely, area above 0.5 $\approx$ 3.1$\times10^{3}$. While RF model yields a much more compact prediction area ($\approx$350), reflecting its strong tendency toward local fitting rather than smooth interpolation, which is clearly visible in the SDM maps. For the phenological datasets (iNat2D and iNat3D), INRs recover a markedly broader and more continuous potential flowering area, whereas RF predictions are confined to the immediate vicinity of observation points. SIREN typically shows the best agreement with observed flowering (highest Hit@1\% and Hit@5\%), Fourier-based models perform slightly worse, and ReLU tends to oversmooth, which leads to the loss of clearly defined centres of biological activity.

\begin{table}[t]
\centering
\scriptsize
\caption{Spatial field summary metrics derived from predicted probability fields.}
\label{tab:eco}
\begin{tabular}{llrrrr}
\toprule
dataset & model & area\_above\_0.5 & EOO (deg$^2$) & Hit@1\% & Hit@5\% \\
\midrule
GBIF   & SIREN   &  3.11e+03 &  31873 & 0.495 & 0.865 \\
GBIF   & FOURIER &  3.11e+03 &  31873 & 0.495 & 0.865 \\
GBIF   & ReLU    &  3.11e+03 &  31873 & 0.495 & 0.865 \\
GBIF   & RF      &  351.7    &  31873 & 0.247 & 0.536 \\
\midrule
iNat2D & SIREN  &  1.15e+03 & 3.219e+04 & 0.346 & 0.722 \\
iNat2D & FOURIER &  1.02e+03 & 3.219e+04 & 0.338 & 0.481 \\
iNat2D & ReLU    &   946.2   & 3.219e+04 & 0.169 & 0.398 \\
iNat2D & RF      &   293.3   & 3.219e+04 & 0.0122 & 0.225 \\
\midrule
iNat3D & SIREN   &  2.72e+03 & 3.219e+04 & 0.297 & 0.662 \\
iNat3D & FOURIER &  2.39e+03 & 3.219e+04 & 0.194 & 0.358 \\
iNat3D & RELU    &  2.51e+03 & 3.219e+04 & 0.125 & 0.463 \\
\bottomrule
\end{tabular}
\end{table}

Table ~\ref{tab:complexity} compares the computational complexity and inference performance of different INR architectures and the RF model. We show the number of parameters, the number of MACs per point, throughput (points/s), and latency (seconds/point). These metrics are important because INR operates on single coordinates and can scale to any grid resolution. Thus, the three INR architectures (SIREN, Fourier, ReLU) have a similar number of parameters ($\approx$ 37,000-50,000) and a very similar number of MACs, with the differences between the datasets stemming only from the input dimensionality. Fourier is the lightest of the INR models $\approx$ 25\% fewer MACs.This translates into slightly higher throughput and lower latency. RF is defined as a baseline with an effective complexity of one parameter, but its throughput does not reflect the true cost of growing trees on large grids. In practice, RF does not scale smoothly to continuous sampling of space or time. 
INR models, by contrast, maintain stable, linear complexity and a predictable computational cost per point. This is crucial in applications that require high-resolution feature reconstruction, such as SDM maps, 3D phenology, or leaf-mask reconstruction.

\begin{table}[t]
\centering
\scriptsize
\caption{Computational complexity and throughput of INR models and the Random Forest baseline.
MACs are counted per forward pass on a single input point; throughput and latency were
measured on a batch of 50\,000 points.}
\label{tab:complexity}
\begin{tabular}{llrrrr}
\toprule
Dataset & Model   & Parameters & MACs & Throughput (pts/s) & Latency (s/pt) \\
\midrule
GBIF   & SIREN    & 50{,}049 & 1.638$\times 10^{7}$ & 1.015$\times 10^{5}$ & 9.86$\times 10^{-6}$ \\
GBIF   & FOURIER  & 37{,}377 & 1.178$\times 10^{7}$ & 1.153$\times 10^{5}$ & 8.67$\times 10^{-6}$ \\
GBIF   & ReLU     & 50{,}049 & 1.638$\times 10^{7}$ & 1.017$\times 10^{5}$ & 9.84$\times 10^{-6}$ \\
GBIF   & RF       &      1   & 0                   & 3.493$\times 10^{5}$ & 2.86$\times 10^{-6}$ \\
\midrule
iNat2D & SIREN    & 50{,}049 & 1.638$\times 10^{7}$ & 9.82$\times 10^{4}$  & 1.02$\times 10^{-5}$ \\
iNat2D & FOURIER  & 37{,}377 & 1.178$\times 10^{7}$ & 1.133$\times 10^{5}$ & 8.83$\times 10^{-6}$ \\
iNat2D & ReLU     & 50{,}049 & 1.638$\times 10^{7}$ & 1.001$\times 10^{5}$ & 1.00$\times 10^{-5}$ \\
iNat2D & RF       &      1   & 0                   & 3.489$\times 10^{5}$ & 2.87$\times 10^{-6}$ \\
\midrule
iNat3D & SIREN    & 50{,}177 & 1.639$\times 10^{7}$ & 9.68$\times 10^{4}$  & 1.03$\times 10^{-5}$ \\
iNat3D & FOURIER  & 37{,}377 & 1.178$\times 10^{7}$ & 1.112$\times 10^{5}$ & 8.99$\times 10^{-6}$ \\
iNat3D & ReLU     & 50{,}177 & 1.639$\times 10^{7}$ & 1.004$\times 10^{5}$ & 9.96$\times 10^{-6}$ \\
\bottomrule
\end{tabular}
\end{table}

Table~\ref{tab:leafsnap} shows the results of leaf mask reconstruction in a point-by-point segmentation/binarized mask on the Leafsnap dataset. Compared to spatial and phenological tasks, this task is much more demanding because it requires the representation of sharp object boundaries and large local variability. The RBF-INR model performs best, achieving the highest F1, IoU, and Dice values, suggesting that radial functions better represent irregular leaf contours than sinusoidal or linear nonlinearities. ReLU comes second, achieving a decent compromise between precision and generalization. Fourier is slightly weaker, likely due to the difficulty in representing very fine structures locally. SIREN performs worst, practically failing to recover masks (F1 $\approx$ 0), indicating that the high oscillation frequency in sinusoidal networks gradually blurs the binary nature of the mask and is not conducive to pixel-wise classification in images with highly irregular edge structure. These results demonstrate that INR architectures can also be applied to morphological tasks, but their effectiveness depends on the type of basis function. RBF models prove to be much more natural for data with sharp edges than SIREN or Fourier networks.

\begin{table}[t]
\centering
\scriptsize
\caption{Leaf morphology segmentation performance on the Leafsnap dataset.
Metrics are computed at the pixel level for the binary leaf/background mask.}
\label{tab:leafsnap}
\begin{tabular}{lrrrrrrr}
\toprule
Model  & LogLoss & Acc  & F1   & Precision & Recall & IoU  & Dice \\
\midrule
ReLU    & 0.276 & 0.874 & 0.340 & 0.552 & 0.245 & 0.205 & 0.340 \\
SIREN   & 0.337 & 0.867 & 0.000 & 0.000 & 0.000 & 0.000 & 0.000 \\
Fourier & 0.281 & 0.872 & 0.310 & 0.545 & 0.217 & 0.184 & 0.310 \\
RBF     & 0.276 & 0.873 & 0.385 & 0.540 & 0.299 & 0.238 & 0.385 \\
\bottomrule
\end{tabular}
\end{table}

In addition to region-based segmentation metrics, we evaluated threshold stability and boundary alignment for the Leafsnap task. For each image, we selected the optimal decision threshold $t^{*}$ on a validation split by maximizing Dice, and compared this per-image thresholding to a single global threshold defined as the median of $t^{*}$ values across images.

Across the evaluated set ($n=46$ images after excluding degenerate masks), per-image threshold selection yielded high region-based performance (mean Dice $0.9721\pm0.0375$, mean IoU $0.9481\pm0.0643$). The optimal thresholds showed moderate variability ($t^{*}$ mean $0.5854\pm0.0993$; median $t_{\mathrm{global}}=0.59$). Importantly, replacing the per-image
thresholds with the single global threshold led to only a minor decrease in area metrics (mean Dice $=0.9685\pm0.0446$, mean IoU $=0.9420\pm0.0730$), indicating that the predicted probability fields are reasonably stable under a fixed operating point.

To quantify contour fidelity, we computed Boundary-F1 under increasing spatial tolerances. Boundary-F1 increased monotonically with tolerance, from $0.265\pm0.182$ at 1\,px, to $0.421\pm0.248$ at 2\,px, $0.615\pm0.281$ at 4\,px, and $0.786\pm0.218$ at 8\,px. This pattern indicates that the implicit field captures the correct large-scale leaf geometry, while strict boundary evaluation penalizes small spatial shifts that are expected for smooth,
continuous reconstructions.

\begin{table}[t]
\centering
\scriptsize
\caption{Boundary-F1 for Leafsnap under increasing spatial tolerance (global threshold $t_{\mathrm{global}}=0.59$). Values are mean $\pm$ SD across $n=46$ images.}
\label{tab:leafsnap_bf1}
\begin{tabular}{lcccc}
\toprule
Tolerance (px) & 1 & 2 & 4 & 8 \\
\midrule
Boundary-F1 & $0.265\pm0.182$ & $0.421\pm0.248$ & $0.615\pm0.281$ & $0.786\pm0.218$ \\
\bottomrule
\end{tabular}
\end{table}

\begin{figure}[t]
\centering
\includegraphics[width=0.75\linewidth]{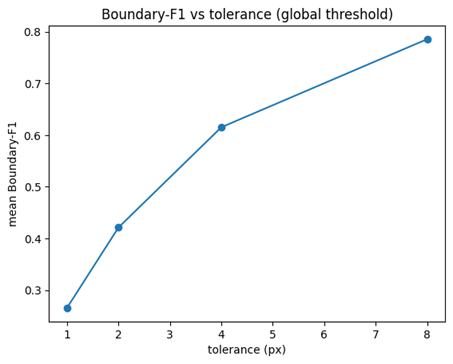}
\caption{Boundary-F1 as a function of spatial tolerance for Leafsnap (global threshold). Boundary agreement increases rapidly with tolerance, indicating accurate large-scale shape reconstruction with small boundary offsets typical of smooth implicit fields.}
\label{fig:leafsnap_bf1_tolerance}
\end{figure}

The Leafsnap analysis highlights a common distinction between region-based and boundary-based evaluation. High Dice/IoU values indicate that the reconstructed leaf area is accurate, whereas Boundary-F1 at strict tolerances penalizes minor contour shifts. The monotonic increase of Boundary-F1 with tolerance suggests that the learned implicit field captures correct global leaf geometry but produces smooth transitions near edges. This behaviour is consistent with
continuous coordinate-based models, which prioritize coherent field reconstruction over pixel-perfect boundary localization.

\begin{table}[t]
\centering
\scriptsize
\caption{Leakage gap quantified as $\Delta = \mathrm{random} - \mathrm{blocked}$ for the GAM-like baseline. Positive $\Delta$ indicates higher (more optimistic) random-split performance; for losses, negative $\Delta$ indicates lower loss under random splits.}
\label{tab:delta_blocked}
\begin{tabular}{lrrrrr}
\toprule
Dataset & $\Delta$ ROC AUC & $\Delta$ PR AUC & $\Delta$ LogLoss & $\Delta$ Brier & $\Delta$ ECE \\
\midrule
GBIF & +0.0158 & +0.0380 & -0.1035 & -0.0366 & -0.0407 \\
iNat2D & +0.0602 & +0.1068 & -0.0775 & -0.0241 & -0.0286 \\
iNat3D & +0.0239 & +0.0770 & -0.0400 & -0.0122 & +0.0031 \\
\bottomrule
\end{tabular}
\end{table}

To summarize leakage inflation, Table~\ref{tab:delta_blocked} reports the gap $\Delta = \mathrm{random} - \mathrm{blocked}$ for the GAM-like baseline.

\begin{figure*}[t]
\centering
\includegraphics[width=\textwidth]{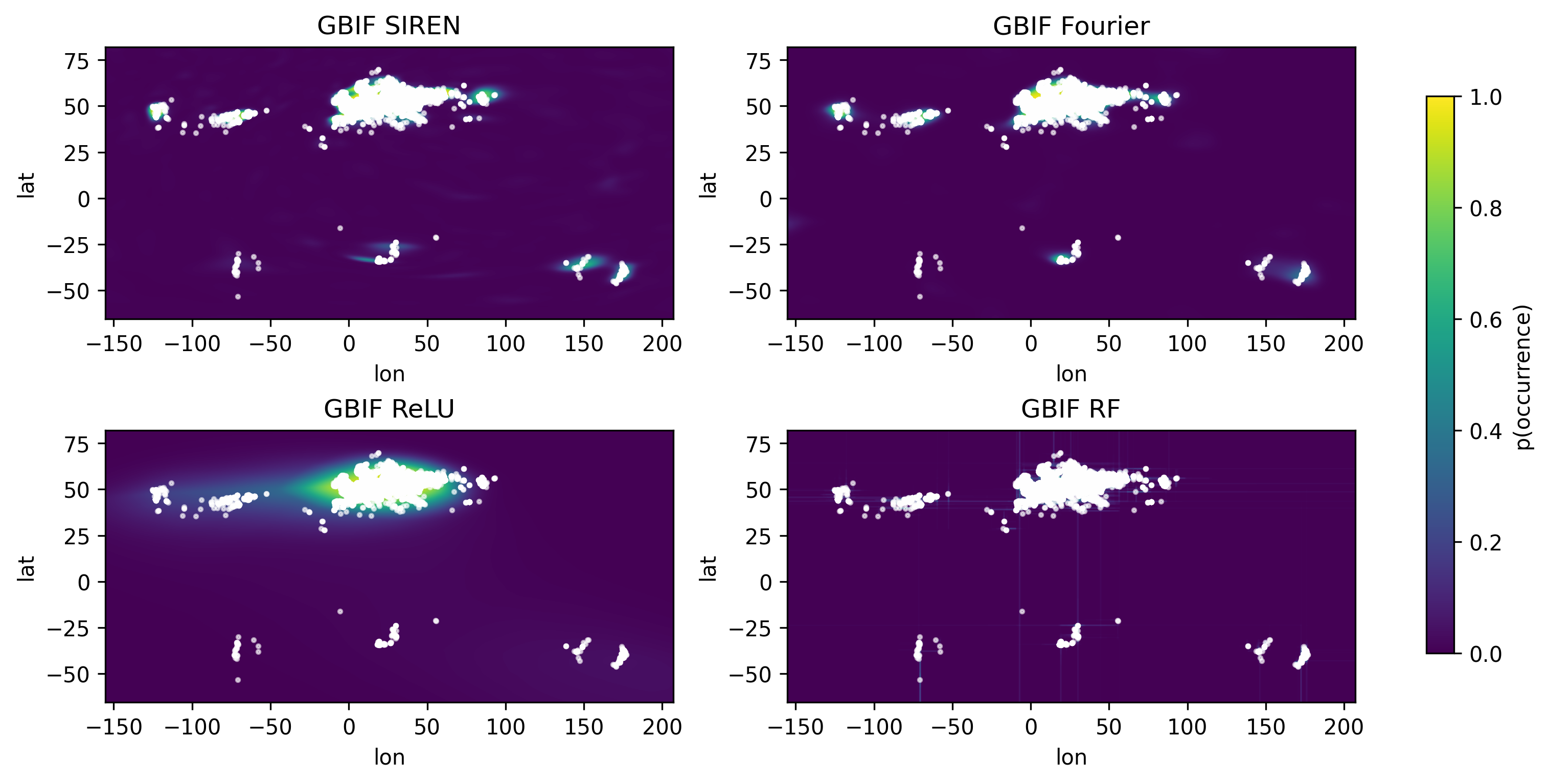}
\caption{SDM for Quercus robur based on GBIF records. Panels compare implicit neural representations (SIREN, Fourier features, ReLU MLP) with a RF baseline. Colours show predicted occurrence probability over a global longitude-latitude grid, white points mark GBIF presences. INR models produce smooth, spatially coherent ranges, whereas the RF exhibits blocky artefacts linked to the training samples.}
\label{fig:gbif_panel}
\end{figure*}

Figure~\ref{fig:gbif_panel} compares four modelling approaches applied to the GBIF occurrence records of Quercus robur. The predictions of the three implicit neural representations: SIREN, Fourier features, and ReLU-MLP are contrasted with RF baseline in a standard longitude-latitude domain. The input consists exclusively of presence-only GBIF data combined with randomly sampled background points, and all models are trained to produce a continuous occurrence probability field on a global grid. Across all INR architectures, the predicted distributions form smooth, spatially coherent envelopes that are consistent with known large-scale biogeographic patterns of a temperate European tree species. The highest probabilities are concentrated across central and western Europe, with gradual decays towards southern and northern limits. Importantly, INRs avoid hard spatial edges and instead represent the species range as a continuous function of geographic coordinates, which is consistent with the expectation of gradual spatial change under continuous environmental gradients, without implying explicit mechanistic drivers. The SIREN model yields the most finely structured patterns, capturing subtle local maxima that correspond to clusters of observations. Fourier features produce slightly more diffuse fields, while ReLU-MLP shows a broader, more smoothed distribution, consistent with its limited ability to represent high-frequency spatial structure. This diversity demonstrates how architectural choices in INR models directly influence spatial expressiveness and interpolation behaviour. In contrast, the Random Forest baseline produces maps dominated by blocky, pixel-like artefacts driven by the discrete partitioning of the geographic space inherent to the algorithm. High-probability regions appear only in the immediate neighbourhood of training points, with sharp edges and empty areas even in regions where environmental continuity would suggest non-zero suitability. Such behaviour is typical for tree ensembles operating on unstructured coordinate data and highlights why classical SDM algorithms require engineered environmental covariates, whereas INR models naturally operate directly in geographic space. Overall, this comparison shows that INR models provide an elegant, coherent way to reconstruct species distributions from unstructured presence-only data, avoiding the artefacts characteristic of tree-based baselines and enabling consistent with known large-scale patterns continuous range representations.

\begin{figure*}[t]
\centering
\includegraphics[width=\textwidth]{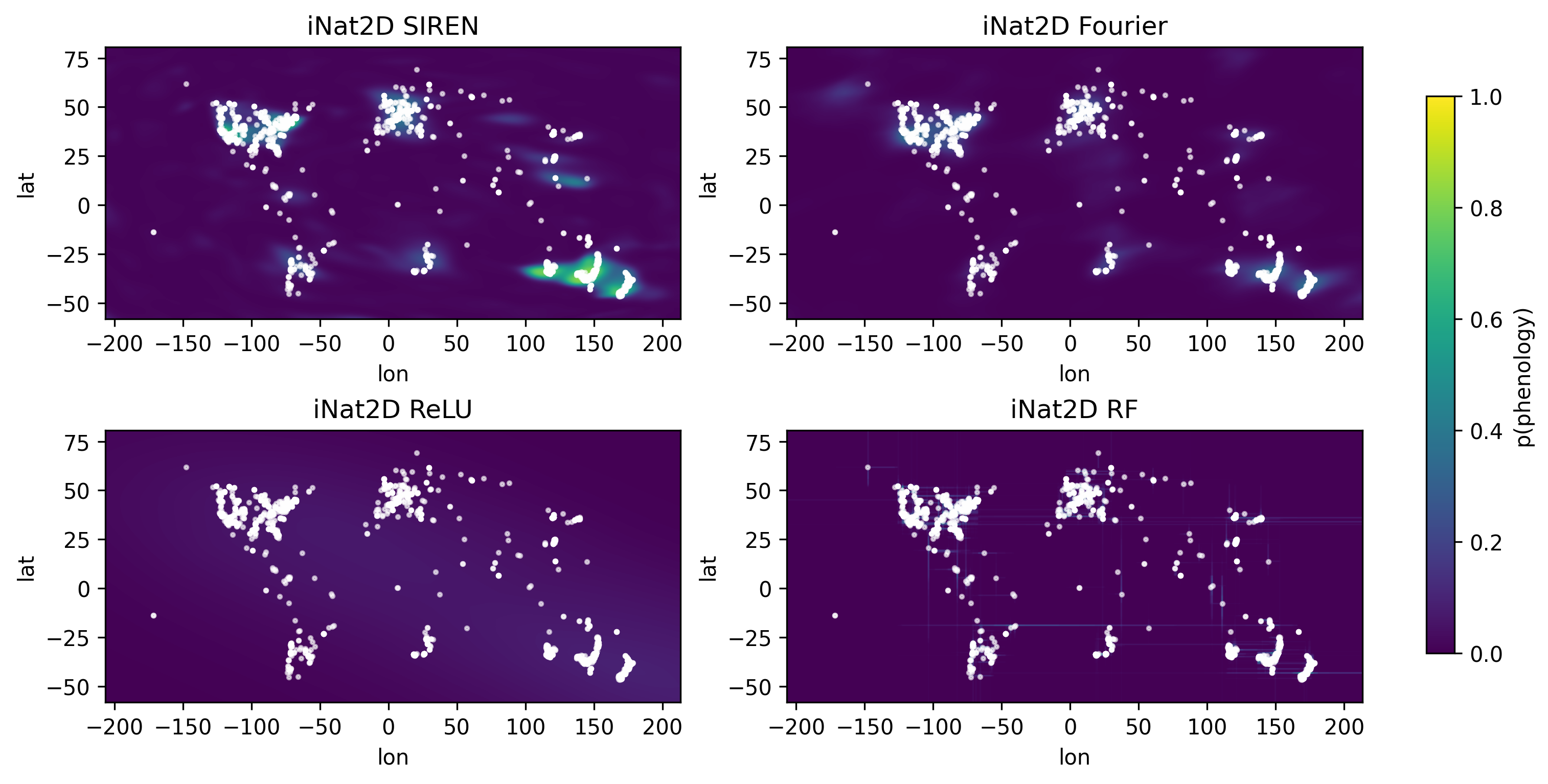}
\caption{
Two-dimensional phenology maps based on iNaturalist flowering observations. Implicit neural representations (SIREN, Fourier features, ReLU MLP) and a RF baseline are trained on presence, background data in longitude-latitude space. Colours represent predicted flowering probability. White points denote iNaturalist observations. INR models reconstruct smooth probability fields, while the RF prediction is highly localized and fragmented around sampling locations.
}
\label{fig:inat2d_panel}
\end{figure*}

Figure~\ref{fig:inat2d_panel} shows spatial flowering-probability maps inferred from iNaturalist observations using the same four models. In this case, the goal is not species distribution but spatial phenology, where presence records correspond to observed flowering events. The task is more challenging because the flowering pattern is shaped at the same time by climate, season, species identity, and sampling bias. Despite these challenges, the INR models again reconstruct continuous, interpretable spatial gradients. For SIREN and Fourier architectures, spatially coherent flowering hotspots emerge in regions known for seasonal peaks in plant observations, particularly western North America, Europe, and eastern Australia. Predictions extend smoothly beyond the observation locations, forming broad probability fields that align qualitatively with global climatic seasonality.  The ReLU-based INR produces even smoother fields with a tendency to over-generalize across continents. This behaviour highlights a key property of INRs: although they interpolate in a continuous manner, the architectural inductive biases strongly shape how sharp or diffuse the resulting spatial maps become. In contrast, RF baseline exhibits extreme spatial fragmentation. High-probability areas are almost exclusively restricted to the exact GPS coordinates of observations, appearing as small isolated pixels. This is an expected limitation of RF trained directly on coordinate features: without auxiliary environmental covariates it cannot generalize beyond the training points. These results show that INRs can be used not only for species range modelling, but also for spatially aggregated ecological processes such as flowering intensity. They do not require environmental predictors or gridded covariates. This can substantially simplify exploratory phenological modelling on global citizen-science platforms.

\begin{figure*}[t]
\centering
\includegraphics[width=\textwidth]{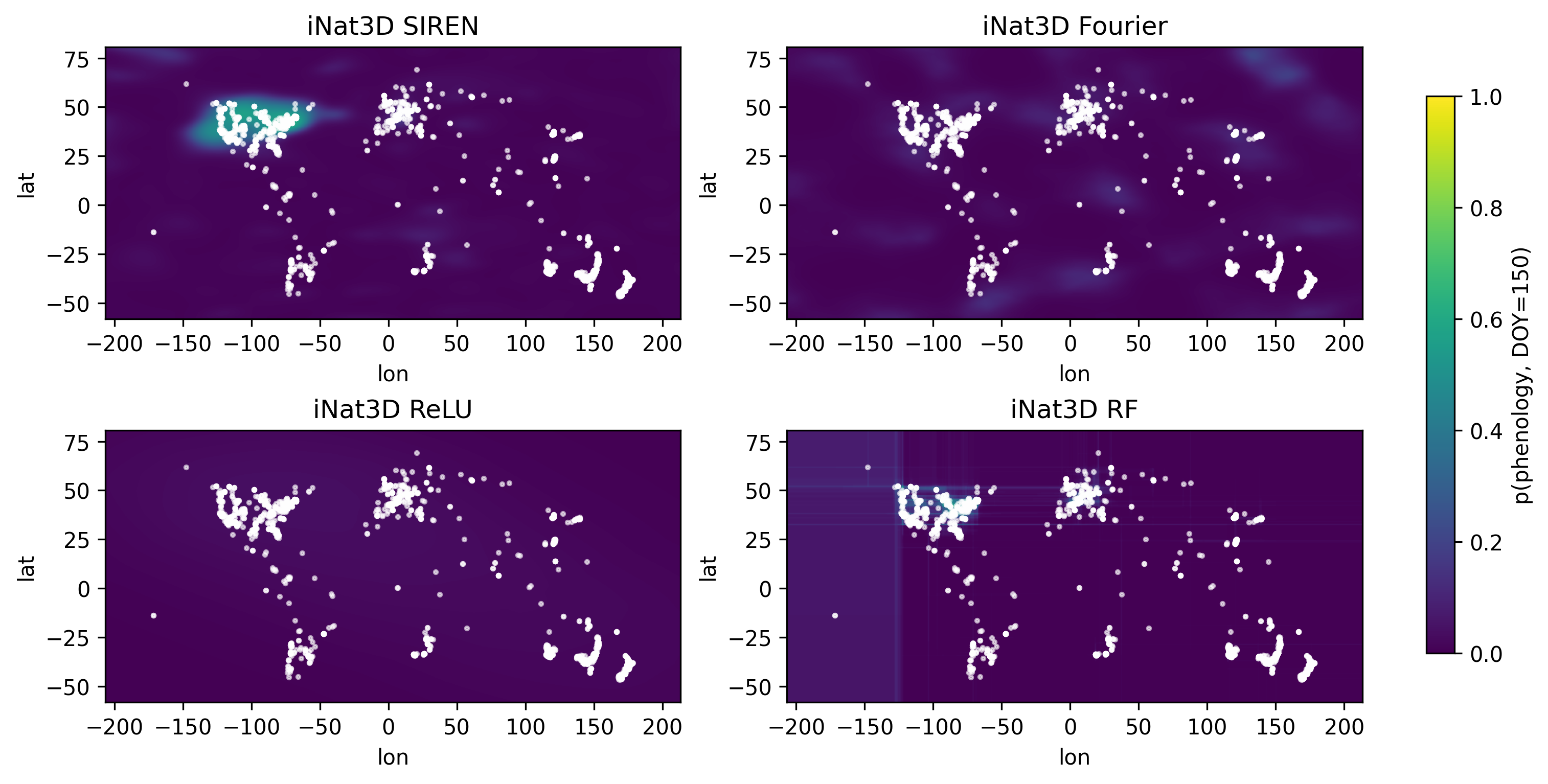}
\caption{Three-dimensional phenology model in lon-lat-DOY  space learned from iNaturalist flowering records. The figure shows a slice at DOY=150 for three implicit neural representations (SIREN, Fourier features, ReLU MLP) and a RF baseline. INR models interpolate flowering probability continuously across space and time, whereas the Random Forest again produces grid-like artefacts tied to the observation pattern.}
\label{fig:inat3d_panel}
\end{figure*}

Figure~\ref{fig:inat3d_panel} shows predictions from models trained on three-dimensional iNaturalist data, where each record includes latitude, longitude, and day-of-year. This task introduces a temporal axis, requiring models to reconstruct phenological dynamics as a smooth function in both space and time. The Figure~\ref{fig:inat3d_panel} presents a slice at DOY = 150 (mid-season in the northern hemisphere), allowing spatial interpretation of flowering probability. The SIREN-based INR yields the most structured spatio-temporal field, identifying coherent phenological hotspots particularly in western North America and central Europe. These regions correspond to well-documented early to mid-season flowering peaks. The sharp yet smooth contours visible in the SIREN map illustrate its ability to represent high-frequency temporal variation through sinusoidal activation functions. The Fourier INR produces broader patterns, with multiple spatial modes corresponding to different regions exhibiting similar seasonal phases. The use of explicit Fourier positional encoding makes this model particularly good at capturing periodicity in DOY, consistent with theoretical expectations. The ReLU-MLP again shows extremely smooth behaviour, predicting large continuous regions of moderate flowering probability. This model’s limited frequency resolution leads to oversmoothing in both space and time, reducing sensitivity to local seasonal variation. The RF baseline performs the weakest in the 3D case. Because tree ensembles partition DOY into discrete bins, the resulting predictions show clear striping effects and discontinuities across the temporal axis. Spatially, the artefacts are even more pronounced than in the 2D case: high-probability regions form rigid, rectangular patches around clusters of observations. This pattern exposes a fundamental limitation of RF for spatio-temporal ecological modelling when using raw coordinates. 
Overall, the 3D results confirm that INRs excel in spatio-temporal interpolation, capturing phenological gradients without requiring environmental covariates, smoothing kernels, or temporal binning. This capacity highlights INRs as a promising tool for reconstructing ecological processes from irregular, opportunistic citizen science data.

\section{Discussion}

From a modelling perspective, implicit neural representations extend the family of continuous environmental models by providing a parametric, resolution-independent representation of spatial and spatio-temporal fields. Unlike grid-based approaches, neural fields decouple model representation from spatial discretisation, which is particularly relevant for large-scale environmental datasets characterised by
heterogeneous sampling density. In this sense, INRs can be interpreted as a modelling layer that complements classical geostatistical methods, generalised additive models, and machine-learning ensembles within environmental software workflows.

Compared to tree-based ensembles trained on raw coordinates, INRs impose a smooth functional 
inductive bias that promotes interpolation rather than memorization. Compared to classical 
smooth spatial models, they offer greater representational flexibility at the cost of calibrated uncertainty. Hybrid approaches that combine INRs with environmental covariates, uncertainty-aware components, or structured spatial priors represent a promising direction for future ecological modelling.

Table~\ref{tab:model_comparison} presents a qualitative comparison of the INR architectures (SIREN, Fourier, ReLU) and the RF baseline. It summarises model behaviour across three categories: predictive performance, ecological realism, and practical aspects.
Stars indicate relative favourability, with more stars corresponding to stronger and more desirable behaviour. INR models achieve smoother, spatially coherent ecological surfaces and robust cross-dataset behaviour, while RF attains artificially perfect AUC due to overfitting to the discrete locations of training points. Fourier features offer the best trade-off between performance and stability, SIREN excels in high-frequency reconstruction, and ReLU is the most stable but produces overly smooth fields. This comparison confirms earlier findings from the INR literature \citep{sitzmann2020,tancik2020} that sinusoidal or frequency-based encodings provide substantially higher representational capacity than standard MLPs. At the same time, the behaviour of the RF model echoes well-known issues of tree-based SDMs trained on coordinate features alone \citep{ElithLaethwick2009,isaac2014}, namely extreme sensitivity to sampling density and inability to interpolate smoothly.

\begin{table}[t]
\centering
\scriptsize
\caption{Qualitative comparison of model behaviour and field reconstruction properties across INR architectures and the RF baseline.}
\label{tab:model_comparison}
\begin{threeparttable}
\begin{tabularx}{\linewidth}{l*{4}{c}}
\toprule
Category & SIREN & Fourier & ReLU & RF \\
\midrule
\textbf{Predictive performance} & & & & \\
Overall ROC/PR AUC
  & ****
  & ****
  & ***
  & *****\tnote{a} \\
Robustness across datasets
  & ****
  & ****
  & ***
  & ** \\
Susceptibility to sampling bias
  & **
  & **
  & ***
  & **** \\
\midrule
\textbf{Spatial field properties} & & & & \\
Spatial smoothness
  & *****
  & ****
  & ***
  & * \\
\midrule
\textbf{Practical aspects} & & & & \\
Parameter count / MACs
  & ***
  & *****
  & ***
  & ***** \\
Training / tuning stability
  & **
  & ****
  & ***
  & ***** \\
\bottomrule
\end{tabularx}

\vspace{4pt}

\begin{tablenotes}
\item[] \textbf{Legend:}\\
\hspace{1em}* = poor / undesirable behaviour\\
\hspace{1em}** = weak\\
\hspace{1em}*** = moderate / acceptable\\
\hspace{1em}**** = strong\\
\hspace{1em}***** = excellent / optimal behaviour
\item[] Categories refer to qualitative properties of reconstructed probability fields (e.g., smoothness, coherence, and behaviour away from sampled locations) rather than to mechanistic ecological interpretability.

\end{tablenotes}

\end{threeparttable}
\end{table}

On the other hand, a comparison of the maps (Figures \ref{fig:gbif_panel}, \ref{fig:inat2d_panel}, \ref{fig:inat3d_panel}) and quantitative metrics shows clear differences between the INR model class and the classical RF. All three INR architectures produce continuous, spatially coherent ecological surfaces, consistent with earlier work showing that coordinate-based networks function as universal function approximators capable of representing smooth and high-frequency signals \citep{sitzmann2020,rahaman2018}. In contrast, the RF model produces highly fragmented distributions, with very high probabilities concentrated exactly at the occurrence sites and pronounced rectangular artefacts caused by the binary partitioning of space by decision trees \cite{ElithLaethwick2009,isaac2014}. Although RF attains artificially high ROC and PR AUC scores, similar overfitting patterns have been reported in SDMs fitted to geographically biased or clustered data \citep{Mendes2020,Broussin2024}. Systematic differences are also evident between the INR models themselves. SIREN is the most high-frequency model, meaning its maps contain fine detail and local maxima, well representing clusters of data, but sometimes also subtle oscillations. This is consistent with the use of sine functions as nonlinearities and the high frequency of the input layer \citep{sitzmann2020,rahaman2018}. The Fourier-feature model still captures complex patterns, but applies stronger smoothing; spatial and temporal structures remain clear, yet are less sharp than in SIREN. ReLU-MLP behaves in the opposite way to SIREN: it produces very smooth, broad fields and often spreads the extent or phenology into large regions of moderate probability. In practice, this reduces sensitivity to high-frequency artefacts, but comes at the cost of weaker localisation of occurrence centres. The differences between models are particularly pronounced in phenological tasks. For iNat2D and iNat3D, where the data distribution is highly skewed, RF effectively reproduces only a point cloud, whereas INRs yield continuous phenology surfaces that support inference in poorly sampled areas. SIREN and Fourier localise flowering hotspots and their spatial extent more precisely, while ReLU produces a broader, global gradient that captures the overall contrast between hemispheres but struggles to separate individual regions. In the 3D setting (lon-lat-DOY), the advantage of INRs over RF is strongest. The tree ensemble treats day of year as just another input variable and introduces sharp transitions between DOY bands, whereas INRs model phenology as a smooth function of time, with seasonal intensity varying within the same regions. From a computational perspective, all three INR architectures are comparable: they have on the order of 4-5$\times 10^{4}$ parameters and a constant complexity per input point. While RF has practically no interpretable equivalent in terms of FLOPs, it is characterized by a larger variability of prediction time depending on the number of trees and their depth. In practice, this means that the qualitative differences between INR and RF are not due to the radically greater complexity of INR models, but mainly to their continuous, coordinate nature.

Implicit neural representations complement, rather than replace, existing ecological modelling approaches. Compared to Random Forests trained on raw coordinates, INRs impose a smooth functional inductive bias that promotes interpolation between observations instead of memorization. Compared to Gaussian process models and spline-based smoothers, INRs replace explicit kernel or basis-function specification with a learnable parametric representation, trading calibrated uncertainty for scalability and expressive power.

Importantly, INRs should be viewed as part of a broader toolkit of continuous machine-learning models for ecological data, alongside Gaussian processes, generalized additive models, Bayesian spatial models, and deep neural networks with environmental covariates. Hybrid formulations that combine INRs with environmental predictors or uncertainty-aware components represent a promising direction for future work.

Citizen-science and occurrence datasets are strongly clustered in space and time.
As a consequence, random train-test splits can substantially overestimate performance due to spatial autocorrelation.
By adding spatially and spatio-temporally blocked protocols, we explicitly quantify this inflation
and show that performance decreases when test points are separated from training regions.
This reinforces that coordinate-only models (including both INRs and classical smoothers)
should be interpreted primarily as interpolative reconstructions unless evaluated under leakage-controlled protocols.

The primary contribution of this study lies in demonstrating how coordinate-based neural fields can serve as a generic layer for continuous environmental field reconstruction from irregular observations. By representing space, time, or image coordinates directly, implicit neural representations decouple field estimation from predefined grids and enable resolution-independent querying. This makes them suitable as preprocessing or representational components within broader environmental modelling pipelines, where reconstructed fields may subsequently be combined with mechanistic models or environmental covariates.

\section{Limitations}
Despite the flexibility and strong empirical performance of implicit neural representations, several limitations of the present study should be acknowledged. First, all models were trained exclusively on presence-background data and did not incorporate environmental covariates. While INRs are capable of learning continuous spatial fields directly from coordinates, ecological mechanisms such as climatic tolerances, soil conditions, and land-use constraints were not explicitly modelled. Classical SDMs typically rely on environmental predictors to capture mechanistic drivers of species distributions \citep{ElithLaethwick2009}, and future INR-based ecological models may benefit from hybrid formulations that integrate mechanistic or covariate-based components. Second, citizen-science datasets such as GBIF and iNaturalist exhibit strong spatial, temporal, and taxonomic biases \citep{isaac2014,cecco2021}. Although INRs interpolate smoothly across irregular sampling patterns, they may still inherit biases present in the training observations. The high predictive scores of the Random Forest baseline demonstrate how machine-learning models can appear highly accurate while being driven primarily by sampling artefacts. INRs mitigate this issue by enforcing spatial and temporal smoothness, yet they do not eliminate biases arising from uneven observer effort or regional differences in reporting.
Third, INRs do not currently provide uncertainty estimates. Traditional SDMs often produce confidence intervals or ensemble-based uncertainty measures. Neural fields, by design, produce deterministic predictions given a fixed set of parameters. Although Bayesian INRs and ensemble variants have been proposed in other domains, they are computationally expensive and we do not consider them here.
As a result, the predicted suitability or phenological intensity is harder to interpret directly in risk-assessment settings. Fourth, although the models scale efficiently with the number of query points, training still requires substantial computational resources.
This is particularly relevant for environmental modelling workflows, where dense gridded observations are often unavailable but resolution-independent predictions are required for exploratory analysis and downstream modelling components. This is particularly the case for SIREN networks, which are sensitive to initialization and frequency scaling \citep{sitzmann2020}.
These factors may hinder adoption in ecological settings where compute resources or machine-learning expertise are limited. Finally, phenological modelling with INRs in this study relied solely on coordinates (latitude, longitude, day-of-year). Phenology is strongly influenced by temperature, photoperiod, and microclimatic variability \citep{millerrushing2008}. Although the continuous nature of INRs allows them to approximate seasonal gradients directly from observations, the absence of environmental drivers limits mechanistic interpretability. 

The continuous probability fields learned by INRs should not be interpreted as mechanistic models of species niches, phenological drivers, or trait evolution. Because no environmental covariates are included, the models cannot explicitly represent physiological tolerances, climatic constraints, or causal ecological processes. Instead, the learned fields reflect smoothed reconstructions of observation intensity or occurrence probability conditional on geographic and temporal coordinates.

Standard INRs used in this study are deterministic and do not provide calibrated uncertainty estimates. Incorporating uncertainty-aware variants (e.g., Bayesian formulations or explicit ensembles with uncertainty maps) is an important direction for future work.

Nevertheless, such reconstructions are valuable as exploratory tools. They can reveal large-scale spatial and temporal structure in heterogeneous biodiversity data, identify undersampled regions, and provide continuous priors for downstream ecological models that incorporate environmental drivers. In this sense, INRs are best understood as descriptive and interpolative models, rather than explanatory or predictive tools for climate-change impact assessment.

Environmental covariates are essential for mechanistic inference and causal interpretation in ecological and environmental models. The coordinate-only formulation adopted here is not intended to replace such approaches, but rather to address an earlier modelling step: continuous field reconstruction under sparse and biased sampling. Integrating implicit neural representations with environmental predictors and uncertainty-aware formulations represents a natural direction for future work.

From an environmental modelling perspective,
implicit neural representations can be interpreted as
a continuous spatial field approximation method,
analogous to spline-based smoothers or Gaussian-process models,
but with different computational scaling properties.

\section{Conclusions}

This study demonstrates that implicit neural representations constitute a practical modelling abstraction for continuous environmental field reconstruction from sparse observations. By representing ecological signals directly as coordinate-based functions, neural fields enable resolution-independent querying and predictable computational scaling, which are desirable properties for environmental modelling software and
large-scale exploratory workflows. Beyond ecological applications, these properties suggest that neural
fields may serve as reusable representation layers within modular environmental modelling software architectures.

Compared to tree-based models trained on raw coordinates, INRs avoid extreme spatial localization and block-like artefacts, yielding reconstructions that better reflect
the gradual nature of ecological variation. Differences between INR architectures highlight the importance of inductive bias: sinusoidal and Fourier-based models are
well suited to capturing fine-scale spatial and temporal structure, whereas simpler architectures provide stronger smoothing. For morphology tasks involving sharp
boundaries, radial basis function variants offer advantages.

We emphasize that the reconstructed fields represent descriptive summaries of observation intensity or relative occurrence conditional on the chosen sampling
design. Because no environmental covariates are included, these models do not provide mechanistic explanations of species niches, phenological drivers, or trait variation.
Nevertheless, such reconstructions are valuable for exploratory analysis, identifying under-sampled regions, and generating continuous spatial priors for downstream
ecological models. In this sense, implicit neural representations should be interpreted primarily as computational tools for spatial field reconstruction rather than as standalone ecological modelling frameworks.

Future work should focus on integrating implicit neural representations with environmental predictors, bias-correction strategies, and uncertainty-aware components to improve interpretability and support ecological decision-making.

\section*{Declaration of competing interests}
The authors declare that they have no known competing financial interests or personal relationships that could have appeared to influence the work reported in this paper.

\section*{CRediT authorship contribution statement}

Agnieszka Pregowska: Conceptualization, Methodology, Software,
Formal analysis, Writing - original draft, Writing - review and editing,
Supervision.

Hazem M. Kalaji: Conceptualization, Formal analysis,
Writing – review and editing.

\clearpage
\appendix
\section*{Supplementary material}
\addcontentsline{toc}{section}{Supplementary Material}

\begin{table}[H]
\centering
\scriptsize
\setlength{\tabcolsep}{3.5pt}
\renewcommand{\arraystretch}{1.15}
\caption{Model architectures, coordinate encodings, and training hyperparameters. Coordinates were normalized to $[-1,1]$ per input dimension.}
\label{tab:s1}
\begin{tabularx}{\linewidth}{l c Y c Y Y}
\toprule
Model & Input dim & Architecture / encoding & Depth$\times$width & Training & Notes \\
\midrule
SIREN
& 2D / 3D
& \makecell[l]{Sine MLP (SIREN)\\$w_0=30$}
& $4\times 128$
& \makecell[l]{Adam\\lr=$10^{-3}$\\batch=4096\\epochs: 10 (2D), 10 (3D)}
& High-frequency, structured fields \\

Fourier-MLP
& 2D / 3D
& \makecell[l]{Random Fourier features\\$K=16$, $\sigma=10$}
& $3\times 128$
& \makecell[l]{Adam\\lr=$10^{-3}$\\batch=4096\\epochs: 10 (2D), 10 (3D)}
& Smooth, stable interpolation \\

ReLU-MLP
& 2D / 3D
& ReLU MLP (no positional encoding)
& $3\times 128$
& \makecell[l]{Adam\\lr=$10^{-3}$\\batch=4096\\epochs: 10 (2D), 10 (3D)}
& Strong smoothing / lower frequency detail \\

RBF-INR
& 2D
& \makecell[l]{Gaussian RBF features\\64 centers, learned $\sigma$}
& $3\times 128$
& \makecell[l]{Adam\\lr=$10^{-3}$\\batch=4096\\epochs: 8}
& Leaf-mask task; sharp boundaries \\
\bottomrule
\end{tabularx}

\vspace{2pt}
\begin{flushleft}
\scriptsize
\textit{Input dim} = coordinate dimensionality. Reported settings extracted from notebooks: \texttt{INR\_eco.ipynb} (2D/3D) and \texttt{INR\_eco\_2.ipynb} (RBF/Leafsnap).
\end{flushleft}
\end{table}

\section*{Software and Data Availability}

\begin{itemize}

\item Name of software: INR-eco

\item Developers: Agnieszka Pregowska

\item Contact: aprego@ippt.pan.pl

\item Date first available: 2025

\item Software required: Python 3.10, PyTorch

\item Program language: Python

\item Source code: upplementary archive accompanying the article. Also a private GitHub repository is available for editorial
and peer-review access:
https://github.com/PregowskaX/INR-eco

The repository will be made publicly available upon publication.

\item Documentation: README file in the repository

\item Example data: toy\_data.csv included in the repository

\item Data sources used in experiments: GBIF, iNaturalist, Leafsnap

\end{itemize}

\section*{Methods}

This section provides algorithmic descriptions of the computational pipeline used in this study. The pseudocode summarizes the main steps of the coordinate-based modelling workflow implemented in the experiments. Algorithm S1 describes the construction of presence-background datasets. Algorithm S2 summarizes the training procedure of implicit neural representations. Algorithm S3 outlines the reconstruction of continuous spatial and spatio-temporal fields. Algorithm S4 describes the evaluation protocols used in the study. Algorithm S5 details the coordinate-based
leaf mask reconstruction procedure used for the Leafsnap dataset.

\begin{algorithm}[t]
\caption{Construction of a coordinate-based presence-background dataset}
\label{alg:dataset}
\begin{algorithmic}[1]
\Require Presence records $P=\{(\mathbf{x}_i, y_i=1)\}_{i=1}^{N_p}$ with coordinates $\mathbf{x}_i \in \mathbb{R}^d$
\Require Number of background samples $N_b$
\Ensure Combined dataset $D=\{(\mathbf{x}_k,y_k)\}_{k=1}^{N}$

\State Remove records with missing or invalid coordinates
\State Remove duplicate coordinate pairs
\State Define spatial (or spatio-temporal) domain $\Omega$ from filtered records
\State Sample background coordinates $\{\tilde{\mathbf{x}}_j\}_{j=1}^{N_b}$ uniformly in $\Omega$
\State Assign labels $y=0$ to all background samples
\State Form the combined dataset
\[
D = \{(\mathbf{x}_i,1)\}_{i=1}^{N_p} \cup \{(\tilde{\mathbf{x}}_j,0)\}_{j=1}^{N_b}
\]
\State Normalize each coordinate dimension to $[-1,1]$
\State Split $D$ into training, validation, and test subsets
\end{algorithmic}
\end{algorithm}

\begin{algorithm}[t]
\caption{Training of a coordinate-based implicit neural representation}
\label{alg:training}
\begin{algorithmic}[1]
\Require Training set $D_{\mathrm{train}}=\{(\mathbf{x}_i,y_i)\}$
\Require Validation set $D_{\mathrm{val}}$
\Require Neural field $f_{\theta}$
\Require Loss function $\mathcal{L}$, learning rate $\eta$, batch size $B$, maximum epochs $T$
\Ensure Trained parameters $\theta^{*}$

\State Initialize network parameters $\theta$
\State Initialize best validation loss $L_{\mathrm{best}} \gets \infty$
\State Initialize patience counter $c \gets 0$
\For{$e=1$ to $T$}
    \State Shuffle $D_{\mathrm{train}}$
    \For{each mini-batch $\mathcal{B} \subset D_{\mathrm{train}}$ of size $B$}
        \State Compute predictions $\hat{y} = f_{\theta}(\mathbf{x})$ for $\mathbf{x} \in \mathcal{B}$
        \State Compute batch loss $\mathcal{L}(\hat{y}, y)$
        \State Update $\theta$ using Adam with learning rate $\eta$
    \EndFor
    \State Evaluate validation loss $L_{\mathrm{val}}$ on $D_{\mathrm{val}}$
    \If{$L_{\mathrm{val}} < L_{\mathrm{best}}$}
        \State $L_{\mathrm{best}} \gets L_{\mathrm{val}}$
        \State Save current parameters $\theta^{*} \gets \theta$
        \State Reset patience counter $c \gets 0$
    \Else
        \State Increment patience counter $c \gets c+1$
    \EndIf
    \If{$c$ exceeds early-stopping patience}
        \State \textbf{break}
    \EndIf
\EndFor
\State \Return $\theta^{*}$
\end{algorithmic}
\end{algorithm}

\begin{algorithm}[t]
\caption{Continuous environmental field reconstruction}
\label{alg:reconstruction}
\begin{algorithmic}[1]
\Require Trained model $f_{\theta^{*}}$
\Require Query domain $\Omega$
\Require Grid resolution $R$
\Ensure Continuous probability field $\hat{p}(\mathbf{x})$

\State Generate a regular query grid $G=\{\mathbf{x}_m\}_{m=1}^{M}$ over $\Omega$ at resolution $R$
\For{each query point $\mathbf{x}_m \in G$}
    \State Compute model output $\hat{p}(\mathbf{x}_m)=\sigma(f_{\theta^{*}}(\mathbf{x}_m))$
\EndFor
\State Reshape predictions on $G$ into a spatial or spatio-temporal field
\State Optionally threshold the field for summary analyses
\State \Return reconstructed field $\hat{p}(\mathbf{x})$
\end{algorithmic}
\end{algorithm}

\begin{algorithm}[t]
\caption{Evaluation under random and blocked protocols}
\label{alg:evaluation}
\begin{algorithmic}[1]
\Require Dataset $D$
\Require Model family $\mathcal{M}$
\Ensure Predictive and spatial summary metrics

\For{each evaluation protocol in \{random, blocked\}}
    \State Partition $D$ into training and test subsets according to the protocol
    \For{each model $m \in \mathcal{M}$}
        \State Train model $m$ on the training subset
        \State Predict on the test subset
        \State Compute classification metrics:
        \Statex \hspace{1.5em} ROC AUC, PR AUC, LogLoss, Brier score, Accuracy@0.5, F1@0.5, ECE
        \State Reconstruct dense probability fields on the query domain
        \State Compute spatial summary metrics:
        \Statex \hspace{1.5em} area above threshold, extent of occurrence, Hit@1\%, Hit@5\%
    \EndFor
\EndFor
\end{algorithmic}
\end{algorithm}

\begin{algorithm}[t]
\caption{Coordinate-based leaf mask reconstruction}
\label{alg:leafsnap}
\begin{algorithmic}[1]
\Require Binary leaf mask image $M(x,y)$
\Ensure Reconstructed probability mask $\hat{M}(x,y)$

\State Sample pixel coordinates $(x_i,y_i)$ from the image plane
\State Assign binary labels $z_i \in \{0,1\}$ from the segmentation mask
\State Normalize image coordinates to $[-1,1]^2$
\State Split sampled pixels into training, validation, and test subsets
\State Train a coordinate-based model $f_{\theta}(x,y)$ using binary cross-entropy
\For{each pixel coordinate on the full image grid}
    \State Compute $\hat{M}(x,y)=\sigma(f_{\theta}(x,y))$
\EndFor
\State Apply threshold $t$ to obtain a binary reconstruction
\State Evaluate Dice, IoU, F1, and Boundary-F1
\end{algorithmic}
\end{algorithm}

\end{document}